\documentclass[letterpaper, 10 pt, conference]{ieeeconf}
\IEEEoverridecommandlockouts
\pdfoutput=1

\usepackage[linesnumbered,vlined,ruled]{algorithm2e}
\usepackage{amsmath}
\usepackage{array}
\usepackage{gensymb}
\usepackage{graphicx}
\usepackage{url}
\usepackage{subcaption}
\usepackage{color}
\usepackage{booktabs}
\usepackage{multicol}

\title{\LARGE \bf Programming by Demonstration with \\ User-Specified Perceptual Landmarks}

\author{Justin Huang and Maya Cakmak%
  \thanks{The authors are with Computer Science and Engineering Department, University of Washington, Seattle, WA 98195, USA.}
}

\begin{document}
  \maketitle
  \thispagestyle{empty}
  \pagestyle{empty}
  
  \begin{abstract}
    Programming by demonstration (PbD) is an effective technique for developing complex robot manipulation tasks, such as opening bottles or using human tools.
    In order for such tasks to generalize to new scenes, the robot needs to be able to perceive objects, object parts, or other task-relevant parts of the scene.
    Previous work has relied on rigid, task-specific perception systems for this purpose.
    This paper presents a flexible and open-ended perception system that lets users specify perceptual ``landmarks'' during the demonstration, by capturing parts of the point cloud from the demonstration scene.
    We present a method for localizing landmarks in new scenes and experimentally evaluate this method in a variety of settings.
    Then, we provide examples where user-specified landmarks are used together with PbD on a PR2 robot to perform several complex manipulation tasks.
    Finally, we present findings from a user evaluation of our landmark specification interface, demonstrating its feasibility as an end-user tool.
  \end{abstract}
  

\section{Introduction}
Programming by Demonstration (PbD) is a user-friendly technique for programming complex manipulation actions on a robot, simply by walking it though the steps of the action.
Previous work has demonstrated that a wide range of object manipulation actions can be represented as a sequence of end-effector poses relative to detected objects or other points of interest ({\em i.e.}, ``landmarks'') in the scene \cite{alexandrova2014robot,niekum2012learning,schulmanlearning}.
However, the expressivity and generalizability of this action representation strongly depends on the robot's ability to robustly perceive landmarks, as well as the granularity of those landmarks.

Unfortunately, previous PbD systems include rigid perceptual systems that limit and pre-specify the types of landmarks that can be referenced in the task.
For instance, one system uses objects detected on a tabletop as landmarks for various manipulation actions \cite{alexandrova2014robot}.
This system will not work for objects on shelves, in clutter, or in scenes that have no horizontal surfaces, such as a doorknob on a door.
In addition, it cannot localize specific {\em parts} of an object.
A different system involves actions relative to object parts as landmarks, {\em i.e.}, ends of a rope to be tied \cite{schulmanlearning}, but these parts are detected with a special-purpose perception system that cannot be used for any other object.
Another system involves easy-to-detect fiducials (also known as Augmented Reality tags) attached to object parts that are to be assembled by the robot \cite{niekum2013incremental}.
This requires attaching fiducials to every single object or scene element that the robot might interact with.

In this paper, we address these limitations by allowing task-relevant landmarks to be specified during the demonstration.
We refer to these as \textit{custom landmarks}.
We present a simple interface for specifying custom landmarks and an algorithm for localizing them in a new scene.
We first characterize the algorithm's ability to localize landmarks in different scenes. 
Then, we demonstrate how manipulation actions programmed on a real robot using custom landmarks can generalize to different scenes.
Finally, through a small-scale user study, we demonstrate the feasibility of our tool for use by novices.
By contributing a flexible and open-ended perceptual system, our work allows manipulation actions that deal with a wide range of objects, object parts, or scene elements in very different environments to be programmed by end-users.


\section{Related work}
This paper presents a perceptual system that expands the set of actions that can be programmed by demonstration.
As such, it builds upon previous work in programming by demonstration, as well as in object localization and point cloud registration.

\subsection{Programming by Demonstration}
PbD is a widely used technique for programming industrial robots, with production-level systems that ship with robots like Kuka's LWR, ABB's Yumi, or ReThink's Baxter, among others. 
It has been an active research area since the 80s, resulting in several surveys on the topic~\cite{billard2008robot,argall2009survey, chernova2014robot}.
While early research explored various representations for encoding and reproducing robot manipulator motions \cite{atkeson1997robot,schaal2005learning,calinon2009statistical,akgun2012trajectories}, recent work has addressed challenges in manipulating objects~\cite{schulmanlearning,alexandrova2014robot,pastor2009learning}, learning high-level task structures~\cite{pardowitz2007incremental,ekvall2008robot,niekum2012learning}, and learning from non-expert demonstrators \cite{akgun2012trajectories,suay2012practical}.
PbD is also closely related to Learning from Demonstration, in which the robot's actions are typically represented as policies in a Reinforcement Learning framework \cite{argall2009survey}.

Previous research on PbD has not focused on scene understanding, instead avoiding the perceptual challenges by using fiducials~\cite{niekum2013incremental}, marker-based motion capture~\cite{wachter2013action}, or simulated environments~\cite{niekum2012learning}.
Others use special-purpose or limited perceptual systems such as object detectors on a flat, uncluttered tabletop surface~\cite{alexandrova2014robot,akgun2016simultaneously}.
In contrast, our work focuses on extending the perceptual component of PbD.
In one work, Ehrennmann et al.~\cite{ehrennmann2000comparison} compared object recognition techniques for PbD.
These techniques were 2D image processing systems that recognized objects with known CAD models available.
In our work, we localize 3D landmarks without assuming that CAD or other models of the landmarks exist.

\subsection{Object localization}
Our paper contributes an algorithm for localizing custom landmarks in new scenes, which builds on the iterative closest point (ICP) algorithm~\cite{besl1992method}.
We represent custom landmarks using a point cloud and attempt to register the point cloud to part of the scene.
Researchers have presented related algorithms to localize objects in a scene.
Rusu et al.\ introduced an algorithm called SAC-IA~\cite{rusu2009fast}, which computes an alignment between two point clouds by repeatedly matching 3D features sampled from the point clouds.
Buch et al.~\cite{buch2013pose} present a similar method with different features and geometric checks to optimize speed.
One way in which our approach differs is that our landmark representation also defines areas of empty space which are expected to be unoccupied around the point cloud.
This allows users to make distinctive landmarks out of otherwise nondistinctive point cloud segments (Fig.~\ref{fig:landmark_box}(a), \ref{fig:landmark_box}(b)).

The problem of localizing landmarks can also be viewed as an object detection and recognition problem, which has seen major progress in recent computer vision research~\cite{anthes2013deep}.
However, these systems are hard to apply to a user-friendly PbD system because they recognize a predefined set of objects and require large amounts of training data to work well.
Additional work has gone into building representations of 3D shapes using deep neural networks~\cite{maturana2015voxnet, wu20153d}.
These representations could be used to compare sampled volumes from the scene to a custom landmark.
However, the current resolution of these representations, $24 \times 24 \times 24$ or $32 \times 32 \times 32$ voxel grids, may be too limited for use in PbD.


\section{Custom landmarks}
\label{sec:custom_landmarks}

\subsection{Programming by Demonstration with landmarks}
\label{subsec:custom_landmarks_repr}

Our system builds on the PbD system by Alexandrova et al., implemented on a dual-arm PR2 robot~\cite{alexandrova2014robot}.
Actions are represented as a sequence of 6D end-effector poses relative to the robot's base or to a {\em landmark}, {\em i.e.,} a point of interest in the environment that can be perceptually detected by the robot.
Each pose also encodes the gripper's state (open or closed). 
Actions are programmed by first asking the robot to detect landmarks in the environments, then kinesthetically moving the robot's arms to desired poses, changing the gripper states if needed, and saving the pose using a verbal command.
A pose becomes relative to a landmark if it is within a certain distance to it; otherwise, it will be relative to the robot's base.
The user can later change the landmark associated with each end-effector pose through a graphical interface.

After programming, the robot executes an action by first localizing all the landmarks in the scene, which may have moved since the demonstration.
The system in \cite{alexandrova2014robot} accomplishes this by performing tabletop segmentation and returning segmented objects as landmarks.
The robot then recomputes end-effector poses defined relative to these landmarks.
Finally, it moves through the poses with both arms, opening or closing its grippers as needed.
An action cannot be executed if a landmark is missing or if a pose is out of reach.

\subsection{Custom landmark representation}
\label{subsec:custom_landmarks_repr}
\begin{figure}
  \centering
  \includegraphics[width=0.99\columnwidth]{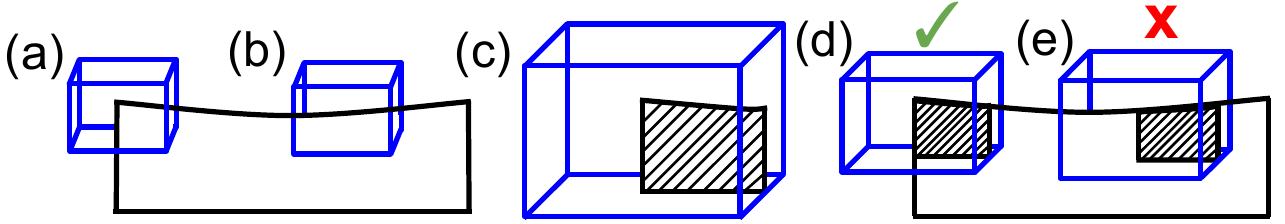}
  \caption{
  An example that illustrates custom landmark specification and search.
    (a) A landmark representing the top left corner of a chair is selected by drawing a box around the corner.
    The top and left halves of the box are empty space.
    (b) A box with empty space above the chair, but not on the sides, represents any part of the top of the chair.
    (c) A close-up view of the landmark captured in (a).
    The shaded region represents the captured point cloud.
    (d) The corner landmark in (c) will match well with the corner of a new chair.
    (e) However, it will not match well when aligned with the top middle of the chair.
    Although all the points in the landmark (shaded region) match well with the chair, some of the chair is inside the rest of the box, which is expected to be unoccupied.
  }
  \label{fig:landmark_box}
\end{figure}

In our system, landmarks are represented by a point cloud that captures the shape of the landmark, as well as a box that surrounds the point cloud.
The boundaries of the box specify margins of empty space around the landmark that are expected to be unoccupied.
The use of empty space is illustrated in Fig.~\ref{fig:landmark_box}.

The user interface for creating landmarks shows a point cloud view from the robot's depth sensor, as well as a 3D box-shaped selector, with controls to set its position and dimensions (see Fig.~\ref{fig:pbd}).
The user moves and resizes the box to surround the point of interest, potentially including margins of empty space on the side.
When the landmark is saved, the system records the subset of the scene within the box, as well as the pose and dimensions of the box.
The interface provides a box-shaped selector aligned with the robot base for simplicity.
However, a more advanced interface could be used to define an arbitrary shape and orientation for the custom landmark.
The landmark is captured from a single view and from a single point cloud, and does not include color information.
Future implementations could fuse multiple point clouds across time or viewpoints.

\subsection{Landmark search algorithm}
\label{sec:system_search_algorithm}
To make use of custom landmarks, we need a way of localizing them in a new scene.
To describe our localization algorithm, we first formally describe the inputs and outputs.
A point $p$ is a location vector $p=(p_x, p_y, p_z)$ and a scene $S$ is simply a set of points.
A landmark $\ell$ is a tuple $(P, B)$, where $P$ is a set of points ({\em i.e.}, points in the point cloud selected by the user) and $B$ represents a box, specified by a 6-dimensional pose and a 3-dimensional size vector.

\begin{algorithm}[t]
  \SetStartEndCondition{ (}{)}{)}%
  \SetAlgoBlockMarkers{}{\}}%
  \SetKwProg{Fn}{}{ \{}{}%
  \SetKwFunction{FindLandmark}{FindLandmark}%
  \SetKwFunction{CandidateError}{CandidateError}%
  \SetKwFor{For}{for}{ \{}{}%
  \SetKwIF{If}{ElseIf}{Else}{if}{\{}{elif}{else\{}{}%
  \SetKwFor{While}{while}{\{}{}%
  \SetKw{KwTo}{in}%
  \SetKwRepeat{Repeat}{repeat\{}{until}%
  \AlgoDisplayBlockMarkers%
  \SetAlgoNoLine%
  \SetKwInOut{Input}{Input}\SetKwInOut{Output}{Output}%
  \Input{\emph{scene}, \emph{landmark}, miscellaneous \emph{parameters}}
  \Output{a list of aligned landmarks in the scene, or empty list if not found}
  crop \emph{scene}\;
  downsample \emph{scene} using a voxel grid\;
  \emph{samples} = randomly sample points from \emph{scene}\;
  \emph{candidates} = []\;
  \For{sample \KwTo samples} {
    \emph{c} = copy of \emph{landmark}\;
    move \emph{c} such that \emph{c.cloud} is centered on \emph{sample}\;
    \emph{c} = run ICP to align \emph{c.cloud} with \emph{scene}\;
    \emph{c.error} = \CandidateError(\emph{c}, \emph{scene})\;
    add \emph{c} to \emph{candidates}\;
  }
  remove all \emph{candidates} that do not have the lowest error within a certain radius (non-max suppression)\;
  \emph{output} = $\{c \in candidates \mid \textit{c.error} < \textit{threshold}\}$\;
  \Return{output}\;
  \caption{FindLandmark}
  \label{alg:find_landmark}
\end{algorithm}

\begin{algorithm}[t]
  \SetStartEndCondition{ (}{)}{)}%
  \SetAlgoBlockMarkers{}{\}}%
  \SetKwProg{Fn}{}{ \{}{}%
  \SetKwFunction{FindLandmark}{FindLandmark}%
  \SetKwFunction{CandidateError}{CandidateError}%
  \SetKwFor{For}{for}{ \{}{}%
  \SetKwIF{If}{ElseIf}{Else}{if}{ \{}{elif}{else\{}{}%
  \SetKwFor{While}{while}{\{}{}%
  \SetKw{KwTo}{in}%
  \SetKw{KwNot}{not}%
  \SetKw{KwContinue}{continue}%
  \SetKwRepeat{Repeat}{repeat\{}{until}%
  \AlgoDisplayBlockMarkers%
  \SetAlgoNoLine%
  \SetKwInOut{Input}{Input}\SetKwInOut{Output}{Output}%
  \Input{\emph{scene}, \emph{candidate} landmark}
  \Output{error score}
  \emph{error} = 0\;
  \emph{denominator} = 0\;
  \emph{croppedScene} = \emph{scene} cropped to \emph{candidate.box}\;
  \emph{visited} = []\;
  \For{scenePt \KwTo croppedScene} {
    \emph{candidatePt} = nearest point in \emph{candidate} to \emph{scenePt}\;
    \emph{error} += distance between \emph{scenePt} and \emph{candidatePt}\;
    \emph{denominator} += 1\;
    add \emph{candidatePt} to \emph{visited}\;
  }
  \For{candidatePt \KwTo candidate.cloud} {
    \If{candidatePt \KwNot\KwTo visited} {
      \emph{scenePt} = nearest point in scene to \emph{candidatePt}\;
      \emph{error} += distance between \emph{scenePt} and \emph{candidatePt}\;
      \emph{denominator} += 1\;
    }
  }
  \Return{error / denominator}\;
  \caption{CandidateError}
 \label{alg:candidate_error}
\end{algorithm}

Our search algorithm takes as input a scene $S$, represented by the complete point cloud captured before an execution, and a landmark $\ell$ to be localized in that scene.
It outputs a set of landmarks $O$, where each landmark $\ell_o \in O$ is potentially an instance of the input landmark $\ell$ in the scene $S$.

Pseudocode for the algorithm is given in Algorithm~\ref{alg:find_landmark}.
First, we crop and downsample the scene according to application parameters.\footnote{
  For this paper, the scene was cropped to a volume in front of the robot roughly equivalent to the reach of its arms.
  The scene was downsampled to a leaf size of 0.005 meters.
  We sampled 5\% of points in the scene, up to a maximum of 1000 points.
  Our non-max suppression radius was 0.03 meters.
  The threshold for our error metric was set to 0.0055 meters.
}
Then, we randomly sample scene points and initialize an instance of $\ell$ at each sampled point.
Next, we run the ICP algorithm to align the landmark's point cloud $P$ with $S$, which produces a modified landmark $\ell'$.
For each landmark $\ell'$, we compute an error metric (Algorithm~\ref{alg:candidate_error}) using $\ell'$ and $S$.
We then perform non-max suppression so that we do not produce duplicates of the same result.
Finally, we filter results by thresholding on the error metric.

The error metric (Algorithm~\ref{alg:candidate_error}) can be thought of as the mean distance between the points of the scene (within the landmark box), and the nearest points of the candidate landmark (lines 5-10), and vice versa (lines 11-17).
Adding a margin of empty space around the landmark helps eliminate false positive matches.
If scene points are found where there is expected to be empty space, then they will increase the mean error, since the nearest points on the landmark are far away.
This process is illustrated in Fig.~\ref{fig:landmark_box}(d) and \ref{fig:landmark_box}(e).

Having all the points of the scene (within the landmark box) match well with the landmark does not imply that all the points of the landmark match well with the scene.
This comes up when the landmark is aligned with a small part of the scene, such that there are more landmark points than scene points.
To avoid false positives in this scenario, we ensure that each point of the landmark contributes to the error score at least once (Algorithm~\ref{alg:candidate_error} lines 11-17).

\subsection{Advantages and limitations}

The perceptual system presented in this paper is flexible enough to represent objects, parts of objects, or parts of the scene.
It has the following advantages, compared to existing perceptual systems for PbD:

\subsubsection{Does not assume a tabletop scene}
Users demarcate landmarks wherever they are in the scene.
A custom landmark can represent an object resting on a table, sitting on a shelf, mounted on a wall, etc.

\subsubsection{Can localize landmarks in hard to segment scenes}
When objects are placed too close together, object segmentation algorithms can easily confuse them for a single object.
The landmark search algorithm, however, searches for custom landmarks uniformly throughout the scene.
This makes it possible to localize a landmark in contact with another object, as long as it remains visible.

\subsubsection{Can represent partly occluded objects}
Even systems that use full object models for landmarks can have a hard time localizing partly occluded objects.
For example, when bowls are stacked, only the rim of the top bowl may be visible, making it hard to match to a model of the bowl.
A custom landmark can be used to represent just the bowl rim, allowing the robot to localize the top bowl of the stack.

\subsubsection{Can generalize across objects with the same part}
Custom landmarks can also be used to represent a part of an object.
For example, while different human tools with similar handles ({\em e.g.}, feather duster, sweeper, squeegee) would normally require a separate object detector for each tool, a custom landmark that represents only the handle could be used to detect all of them and easily transfer actions demonstrated for one tool to the the other.

\subsubsection{Does not require task-specific detectors}
Our system localizes custom landmarks based on their shape and is agnostic to the task at hand.
This allows it to be used for a variety of tasks that might otherwise require custom perception systems ({\em e.g.}, handle detectors, bowl or bowl rim detectors).
This is also useful for fixed parts of the scene.
For example, a custom landmark can be used to localize some prominent feature of a laundry control panel, such as the central dial, or a corner of the machine.
The robot can then press different buttons on the panel, whose positions are known as offsets from the landmark.

\subsubsection{Does not assume ``objectness''}
Custom landmarks can be arbitrary parts of a scene, such as a drawer handle, the top slot of a recycling bin, or a window sill.
They can even represent the lack of objects, {\em e.g.,} using a flat horizontal patch to search for an empty spot on a cluttered surface.

The main limitation of our system is that it relies on landmarks having a unique shape in the scene.
In semi-structured environments, it could be the case that the scene is designed to not distract from the landmarks, but this is not true in general.
Custom landmarks also do not make use of color information, so they can't distinguish between different colors of the same object, or recognize color patterns.
Because we only capture custom landmarks from a single view of a point cloud, it can be brittle to viewpoint changes, such as when a non-symmetric object is rotated.
Finally, our system puts the burden on users to intelligently create the custom landmarks to maximize their uniqueness.
To provide good landmarks, users must understand how Algorithm~\ref{alg:find_landmark} and \ref{alg:candidate_error} work, which requires explanation and training.


\section{Search algorithm experiments}
Our first evaluation characterizes the search algorithm's ability to localize landmarks in a variety of settings.

\subsection{Settings}
\label{sec:settings}

\begin{figure*}
  \centering
  \includegraphics[width=\textwidth]{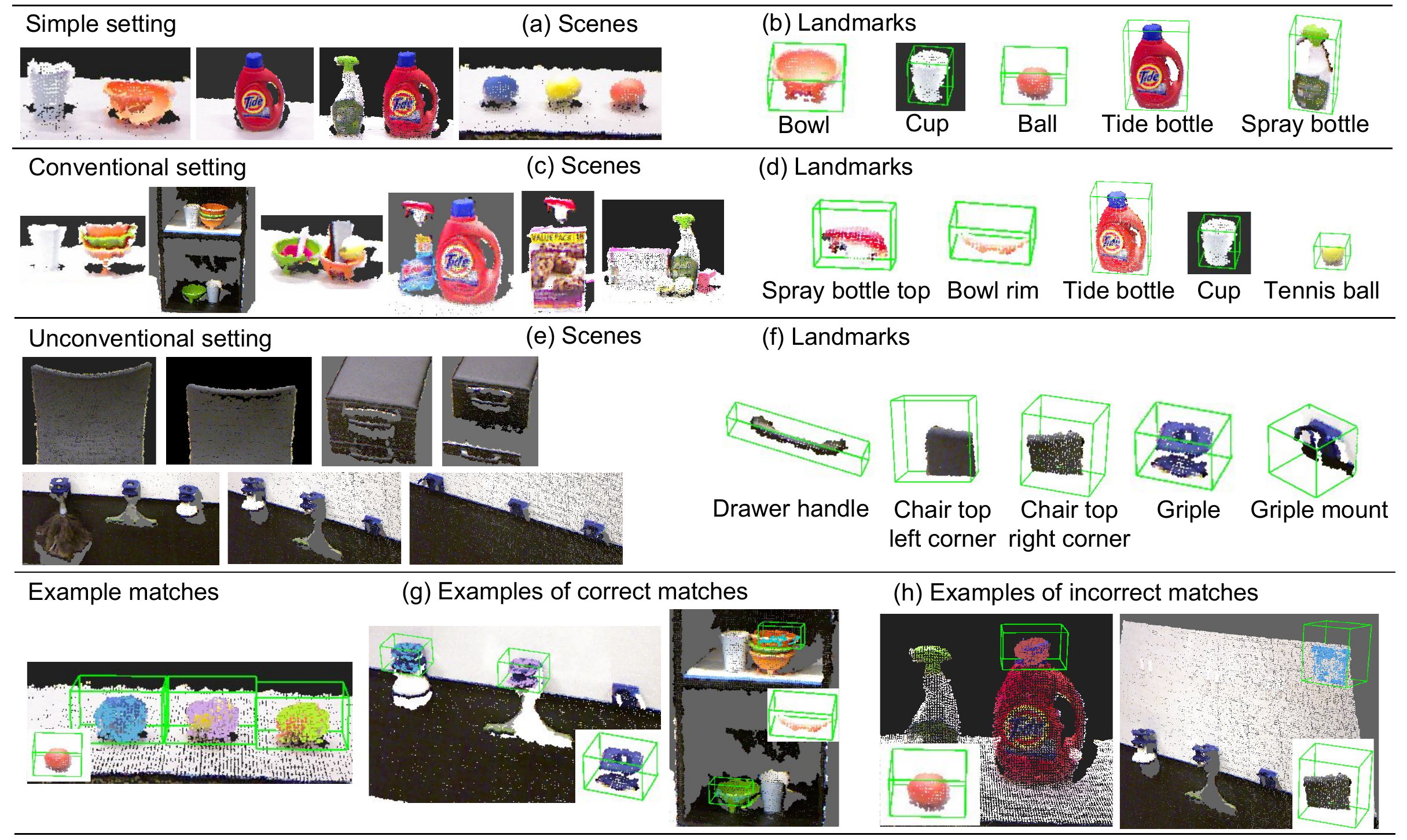}
  \caption{
    (a) Scenes in the \textit{simple} setting were such that objects could be easily segmented.
    (b) Landmarks in the \textit{simple} setting all represented whole objects.
    (c) Scenes in the \textit{cluttered} setting had objects placed in contact or stacked on each other. This setting also included a shelf scene.
    (d) Some landmarks in the \textit{cluttered} setting represented parts of objects, such as the top of a spray bottle or the rim of a bowl.
    (e) Scenes in the \textit{unconventional} setting were not on tabletops.
    (f) Landmarks in the \textit{unconventional} setting were unique objects like Griples or Griple mounts. Or, they were part of the scene, such as a chair corner or drawer handle.
    (g) Our system worked in simple scenes. It also found the two bowl rims on the top and bottom shelves of the shelf scene, and found unique landmarks like Griples.
    (h) Our system incorrectly confused the cap of a Tide bottle with a ball, and found a chair corner at the corner of a vertical surface.
  }
  \label{fig:dataset}
\end{figure*}

The algorithm was evaluated in three kinds of settings: \textit{simple}, \textit{cluttered}, and \textit{unconventional}.
Different sets of landmarks were created for each setting (Fig.~\ref{fig:dataset}).

In the \textit{simple} setting (Fig.~\ref{fig:dataset}(a), \ref{fig:dataset}(b)), objects were spaced apart on a tabletop and the landmarks represented whole objects.
These scenes, detailed in Table~\ref{tab:evaluation_simple_scenes}, were representative of scenes used in previous PbD work and showed how our system could function in lieu of tabletop segmentation.

\renewcommand{\arraystretch}{1.25}
\begin{table}
  \caption{Scenes and landmarks in the \textit{simple} setting.}
  \label{tab:evaluation_simple_scenes}
  \centering
  \begin{tabular}{m{0.5\columnwidth} m{0.35\columnwidth}}
    \hline
    \small{\textbf{Scene}}                   & \small{\textbf{Landmarks}} \\
    \hline
    \small{A bowl and a cup on a table}      & \small{Bowl, cup} \\
    \small{A Tide bottle on a table}         & \small{Tide bottle} \\
    \small{A Tide bottle and a spray bottle} & \small{Tide bottle, spray bottle} \\
    \small{Three balls on a table}    & \small{Balls} \\
    \hline
  \end{tabular}
\end{table}

In the \textit{cluttered} setting, detailed in Table~\ref{tab:evaluation_conventional_scenes} and illustrated in Fig.~\ref{fig:dataset}(c) and \ref{fig:dataset}(d), objects were in contact or occluding one another.
Additionally, one of the scenes was a shelf scene, rather than a tabletop.
Some of the landmarks represented parts of objects rather than the whole object itself.
These scenes were designed such that tabletop segmentation would not have worked.

\renewcommand{\arraystretch}{1.5}
\begin{table}
  \caption{Scenes / landmarks in the \textit{cluttered} setting.}
  \label{tab:evaluation_conventional_scenes}
  \centering
  \begin{tabular}{m{0.59\columnwidth} m{0.31\columnwidth}}
    \hline
    \small{\textbf{Scene}}                                                           & \small{\textbf{Landmarks}} \\
    \hline
    \small{A stack of bowls and a cup on a table}                                    & \small{Rim of top bowl, cup} \\
    \small{Two cups and stacked bowls placed on two different levels of a shelf}     & \small{Rims of top bowls, cups} \\
    \small{Bowls placed in contact, containing items including a tennis ball, on a table}           & \small{Rims of bowls,\newline tennis ball} \\
    \small{A Tide and a spray bottle in contact on a table}                          & \small{Tide bottle,\newline top of spray bottle} \\
    \small{A spray bottle, occluded on bottom}                            & \small{Top of spray bottle} \\
    \small{A different spray bottle with tennis balls and other clutter, on a table} & \small{Top of spray bottle,\newline tennis balls} \\
    \hline
  \end{tabular}
\end{table}

The \textit{unconventional} setting (Fig.~\ref{fig:dataset}(e), \ref{fig:dataset}(f), Table~\ref{tab:evaluation_unconventional_scenes}) did not contain any tabletop scenes, and landmarks represented parts of the scene ({\em e.g.}, a drawer handle) or unique objects like Griples.\footnote{
  Griples are 3D-printed adaptors that can be attached to tool handles to make them easier for the PR2 to grasp, see~\cite{xu2014enhanced}.
  Tools with Griples can be hung from 3D-printed Griple mounts (Fig.~\ref{fig:dataset}(f)) fastened to a wall.
}
This was designed to demonstrate how custom landmarks can handle recognition tasks that might otherwise require a custom, task-specific detectors.

\renewcommand{\arraystretch}{1.5}
\begin{table}
  \caption{\textit{Unconventional} scenes and landmarks.}
  \label{tab:evaluation_unconventional_scenes}
  \centering
  \begin{tabular}{m{0.5\columnwidth} m{0.4\columnwidth}}
    \hline
    \small{\textbf{Scene}}                                  & \small{\textbf{Landmarks}} \\
    \hline
    \small{Chair in front of robot}                         & \small{Corners of chair backrest} \\
    \small{Lowered chair in front of robot}                 & \small{Corners of chair backrest} \\
    \small{Partly full Griple tool rack} & \small{Griples, Griple mounts} \\
    \small{Full Griple tool rack}                           & \small{Griples, Griple mounts} \\
    \small{Empty Griple tool rack}                          & \small{Griples, Griple mounts} \\
    \small{An under-desk drawer}           & \small{Drawer handles} \\
    \small{The under-desk drawer with\newline bottom drawer open}   & \small{Drawer handles} \\
    \hline
  \end{tabular}
\end{table}

\subsection{Measures}
\label{subsubsec:evaluation_measures}
The localization algorithm was run for each scene and landmark pair per setting.
The correctness of the output was hand-labelled by the first author.
We report the precision and recall of the algorithm under three different assumptions:
\begin{enumerate}
  \item All landmarks are searched for in all scenes.
  \item Only certain landmarks are searched for in certain scenes ({\em i.e.}, they are \textit{in context}), as enumerated in Tables~\ref{tab:evaluation_simple_scenes}, \ref{tab:evaluation_conventional_scenes}, and \ref{tab:evaluation_unconventional_scenes}.
    This can be a valid assumption in real-world settings.
    For example, if the robot goes to a Griple tool rack, it is probably either searching for a Griple or a Griple mount, but not a chair.
  \item In addition to assumption 2, if multiple instances of a landmark are in a scene, we only want to localize one of them.
    This is a valid assumption for PbD actions, which only operate on one landmark at a time. 
\end{enumerate}

\subsection{Results}
\label{sec:eval_results}

Precision and recall values for the evaluation settings described above are summarized in Table~\ref{tab:evaluation_scores}.
The results show that our system works better in situations where landmarks are searched for {\em in context}.
The recall rate is further increased when the system only needs to find one instance of a landmark---this shows that the system will often find at least one instance of a landmark, even though it may not find them all.
Our system is least successful when searching for arbitrary objects in arbitrary scenes, which results in a larger number of false positives.

\renewcommand{\arraystretch}{1.25}
\begin{table}
  \caption{
    Precision / recall scores of the landmark search algorithm, under the assumptions from Section~\ref{subsubsec:evaluation_measures}.
  }
  \label{tab:evaluation_scores}
  \centering
  \begin{tabular}{m{0.6\columnwidth} m{0.12\columnwidth} m{0.1\columnwidth}}
    \toprule
    \small{\textbf{\textit{Simple} setting assumption}}                & \small{\textbf{Precision}} & \small{\textbf{Recall}} \\
    \hline
    \small{All landmarks searched in all scenes} & \small{72.72\%} & \small{100\%} \\
    \small{Landmarks and scenes in context}   & \small{100\%}   & \small{100\%} \\
    \small{Only find one instance of landmark}     & \small{100\%}   & \small{100\%} \\
    \midrule
    \hline
    \small{\textbf{\textit{Cluttered} setting assumption}}                & \small{\textbf{Precision}} & \small{\textbf{Recall}} \\
    \hline
    \small{All landmarks searched in all scenes} & \small{75\%}  & \small{60\%} \\
    \small{Landmarks and scenes in context}   & \small{100\%} & \small{60\%} \\
    \small{Only find one instance of landmark}     & \small{100\%} & \small{81.82\%} \\
    \midrule
    \hline
    \small{\textbf{\textit{Unconventional} setting assumption}}                      & \small{\textbf{Precision}} & \small{\textbf{Recall}} \\
    \hline
    \small{All landmarks searched in all scenes} & \small{64.71\%} & \small{64.71\%} \\
    \small{Landmarks and scenes in context}   & \small{100\%} & \small{64.71\%} \\
    \small{Only find one instance of landmark}     & \small{100\%} & \small{80\%} \\
    \bottomrule
  \end{tabular}
\end{table}

In the \textit{simple} setting, our system mistakenly localized a ball at the cap of a Tide bottle.
This is shown in Fig.~\ref{fig:dataset}(h).
This error occurred in both scenes which had the Tide bottle in it.
The system localized all other objects in all other scenes in this setting correctly.

In the \textit{cluttered} setting, our system had the same issue with a tennis ball being localized at the cap of a Tide bottle.
It also mislocalized the spray bottle top to the back of the bottom shelf in the shelf scene.
The back of the shelf had an abnormal appearance due to the robot not having full depth data.
This shows that, in our system, the smaller and less constrained the landmark, the more likely it is to be matched incorrectly in the scene.
However, the system was able to successfully localize objects in cluttered and partly occluded configurations.
For example, it was able to localize a bowl rim in a stack of bowls, including in the shelf scene (Fig.~\ref{fig:dataset}(g)).
In the scene where a spray bottle was in contact with a Tide bottle (Fig.~\ref{fig:dataset}(c)), the system was able to localize both bottles individually, whereas a tabletop segmentation algorithm would consider them to be one large object.

In terms of recall, the system failed to find some expected landmarks.
In one scene, we put a tennis ball inside a bowl, to see if the system would find either the tennis ball or the bowl rim.
Both objects intruded on the expected empty space regions of the other, so the system found neither.

In the \textit{unconventional} setting, the only false positive was to localize the corners of a chair at the corners of a whiteboard, from which we hung Griples.
This error was repeated and counted multiple times in our data, for the left and right chair corners, and for all three scenes using the Griple tools.
It also failed to find the drawer handles in one of the two drawer scenes.
However, the system did manage to localize objects like Griples, Griple mounts, and chair corners in the appropriate scenes.

\subsection{Discussion}
While our results demonstrated that our perception system can perform what it was designed to do as part of a PbD system, it was not robust in the general perception problem of finding arbitrary objects in arbitrary scenes.
In particular, small objects are more likely match unexpected parts of the scene ({\em e.g.}, the ball being matched to the cap of the Tide bottle).
Distinctive landmarks, such as those with empty space in its interior, work better.
For example, although the bowl rim landmark is small, our error metric will not confuse it with most areas of the scene, because it expects to find empty space inside bowl rim's proposed location.
Inversely, the chair backseat corners, which had little empty space inside, were confused with the corners of the whiteboard.
Although the intent of the chair corner landmark was to only match with other chairs, it will in fact match with the corner of any flat surface.
This suggests that users should try to design custom landmarks to be unique-looking.
For example, if it can be assumed that balls are on a tabletop, then it could help to include part of the tabletop beneath the ball as part of the landmark.
This would avoid the ball being confused with the similarly-shaped Tide bottle cap.

\renewcommand{\arraystretch}{1.25}
\begin{table}[t]
  \caption{\label{tab:pbd_eval1} Demonstration Task 1: Hanging tools}
  \begin{tabular}{m{0.8\columnwidth}  m{0.1\columnwidth}}
    \midrule
    \multicolumn{2}{m{0.95\columnwidth}}{\small{\textbf{Scene and objects:} Three Griple mounts on a vertical surface and three tools with Griples that could hang from the mounts}}\\
    \multicolumn{2}{m{0.95\columnwidth}}{\small{\textbf{Landmarks:} Griple, Griple mount}}\\
    \multicolumn{2}{m{0.95\columnwidth}}{\small{\textbf{Actions:} Picking up tools from the rack and hanging them back}}\\
    \midrule
    \small{\textbf{Test scenarios}} & {\em \textbf{Success?}} \\
    \small{Picking the only tool from a rack (Fig.~\ref{fig:pbd}(a))} & {\em Yes} \\
    \small{Picking a tool from a full rack} & {\em Yes} \\
    \small{Picking a tool with no tools available (should not act)} & {\em Yes} \\
    \small{Placing a tool on an empty rack} & {\em Partial} \\
    \small{Placing a tool on an full rack (should not act)} & {\em Yes} \\
    \midrule
  \end{tabular}
\end{table}

\renewcommand{\arraystretch}{1.25}
\begin{table}
  \caption{\label{tab:pbd_eval2} Demonstration Task 2: Grasping a chair} 
  \centering
  \begin{tabular}{m{0.8\columnwidth}  m{0.1\columnwidth}}
    \midrule
    \multicolumn{2}{m{0.95\columnwidth}}{\small{\textbf{Scene and objects:} The back of an office chair facing the robot}}\\
    \multicolumn{2}{m{0.95\columnwidth}}{\small{\textbf{Landmarks:} Top of the chair backrest}}\\
    \multicolumn{2}{m{0.95\columnwidth}}{\small{\textbf{Actions:} Grasp the backrest of the chair with two arms}}\\
    \midrule
    \small{\textbf{Test scenarios}} & {\em \textbf{Success?}} \\
    \small{Five different scenes with the chair raised to different heights, placed in different positions and orientations} & {\em 4/5} \\
    \bottomrule
  \end{tabular}
\end{table}

\renewcommand{\arraystretch}{1.25}
\begin{table}
  \caption{\label{tab:pbd_eval3} Demo Task 3: Picking bowls from a shelf} 
  \centering
  \begin{tabular}{m{0.8\columnwidth}  m{0.1\columnwidth}}
    \midrule
    \multicolumn{2}{m{0.95\columnwidth}}{\small{\textbf{Scene and objects:} Similar to the shelf scene in the \textit{cluttered} setting (Section \ref{sec:settings}), with bowls in contact with a paper cup}. Scene-1: one bowl on the top and one bowl on the bottom shelf; Scene-2: two bowls stacked on the top shelf; Scene-3: only cups.}\\
    \multicolumn{2}{m{0.95\columnwidth}}{\small{\textbf{Landmarks:} Bowl rim}}\\
    \multicolumn{2}{m{0.95\columnwidth}}{\small{\textbf{Actions:} Picking up a bowl from its rim}}\\
    \midrule
    \small{\textbf{Test scenarios}} & {\em \textbf{Success?}} \\
    \small{Picking the two bowls in scene-1} & {\em 2/2} \\
    \small{Picking the two bowls in scene-2 (Fig.~\ref{fig:pbd}(c))} & {\em 2/2} \\
    \small{Picking up a bowl in scene-3 (should not act)} & {\em Yes} \\
    \bottomrule
  \end{tabular}
\end{table}

\renewcommand{\arraystretch}{1.25}
\begin{table}
  \caption{\label{tab:pbd_eval4}Demo Task 4: Trash bin / recycling bin}
  \centering
  \begin{tabular}{m{0.8\columnwidth}  m{0.1\columnwidth}}
    \midrule
    \multicolumn{2}{m{0.95\columnwidth}}{\small{\textbf{Scene and objects:} Trash and recycling bins in front of the robot; both bins have same dimensions, but the trash bin has a square opening, while the recycling bin has a circular opening.}}\\
    \multicolumn{2}{m{0.95\columnwidth}}{\small{\textbf{Landmarks:} Square and circular openings of the bins}}\\
    \multicolumn{2}{m{0.95\columnwidth}}{\small{\textbf{Actions:} Drop an item into either bin}}\\
    \midrule
    \small{\textbf{Test scenarios}} & {\em \textbf{Success?}} \\
    \small{Drop in trash bin, with both bins moved slightly} & {\em Yes} \\
    \small{Drop in trash bin, with two bins swapped (Fig.~\ref{fig:pbd}(d))} & {\em Yes} \\
    \small{Drop in recycling bin, with two bins rotated $45^o$} & {\em Yes} \\
    \small{Drop in trash bin, with only recycling bin in the scene} & {\em Yes} \\
    \small{Drop in recycling bin, only recycling bin in the scene} & {\em Yes} \\
    \bottomrule
  \end{tabular}
\end{table}

\section{Demonstrations on a PR2 robot}
\label{subsec:system_demos}
\begin{figure*}[t]
  \centering
  \includegraphics[width=\textwidth]{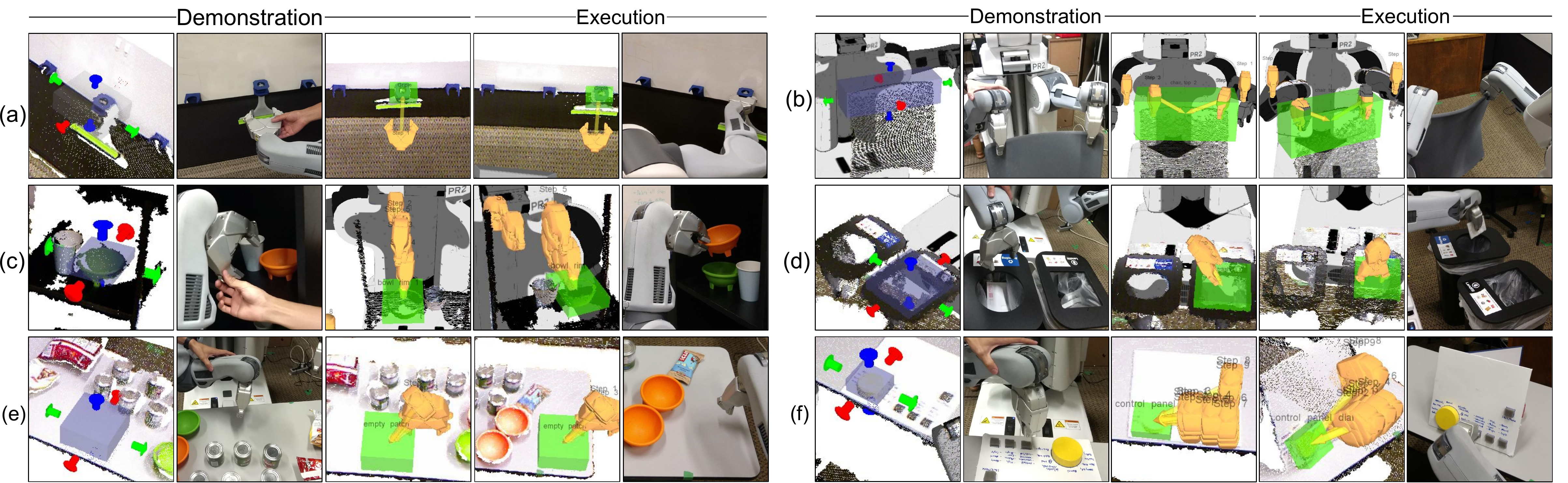}
  \caption{
    PbD demonstrations with custom landmarks, as described in Section~\ref{subsec:system_demos}.
    Each row (a)-(f) illustrates Tasks 1-6, respectively.
    The pictures in each row show, from left to right,
    1) the box the user defines around the custom landmark of interest,
    2) the user demonstrating an action relative to that landmark,
    3) a visualization of the demonstrated action (orange grippers), and the custom landmark (green box),
    4) a visualization of the PbD execution in a different scene, and
    5) the robot executing the action in the new scene.
    (a) Shows the robot picking a Griple tool, which was moved to a different position in one test.
    (b) Shows the robot grasping a slightly rotated chair.
    (c) Shows the robot picking the top bowl from a stack of bowls on a shelf.
    (d) Shows the robot placing an item in either a trash bin or a recycling bin.
    (e) Shows the robot finding an empty patch on a crowded table.
    (f) Shows the robot operating a mockup of a laundry control panel.
    During the demonstration, the control panel was flat on the table, but during one test, the control panel was upright.
  }
  \label{fig:pbd}
\end{figure*}

This section provides several examples of how custom landmarks can be put into practice within PbD.
We describe six complex manipulation actions that we programmed by demonstration using custom landmarks. 
These actions were not possible using previous systems such as in \cite{alexandrova2014robot}.
An important aspect of these tasks was that they were programmed without writing any custom perception or manipulation code, or doing any other technical work like running scripts.
Each action was programmed and executed in five new scenes, different from the demonstration scene.
Performing the tasks, including providing the demonstration and defining custom landmarks, took less than 30 minutes to do, each.
The scenes, landmarks, programmed actions, and the test scenarios for each task are explained in Tables~\ref{tab:pbd_eval1}-\ref{tab:pbd_eval6}.

\renewcommand{\arraystretch}{1.25}
\begin{table}[t]
  \caption{\label{tab:pbd_eval5} Demonstration Task 5: Crowded surface}
  \centering
  \begin{tabular}{m{0.8\columnwidth}  m{0.1\columnwidth}}
    \midrule
    \multicolumn{2}{m{0.95\columnwidth}}{\small{\textbf{Scene and objects:} The robot searches for an empty spot on a crowded tabletop with various objects}}\\
    \multicolumn{2}{m{0.95\columnwidth}}{\small{\textbf{Landmarks:} A flat tabletop patch}}\\
    \multicolumn{2}{m{0.95\columnwidth}}{\small{\textbf{Actions:} Point to an empty spot}}\\
    \midrule
    \small{\textbf{Test scenarios}} & {\em \textbf{Success?}} \\
    \small{Different scenes with objects configured to vary the position of the empty patch. Two of the scenes had a box larger than the empty patch on the table.} & {\em 4/5} \\
    \bottomrule
  \end{tabular}
\end{table}

\renewcommand{\arraystretch}{1.25}
\begin{table}
  \caption{\label{tab:pbd_eval6}Task 6: Control panel}
  \centering
  \begin{tabular}{m{0.8\columnwidth}  m{0.1\columnwidth}}
    \midrule
    \multicolumn{2}{m{0.95\columnwidth}}{\small{\textbf{Scene and objects:} Control panel, similar to a laundry machine, with push buttons and a large central dial}}\\
    \multicolumn{2}{m{0.95\columnwidth}}{\small{\textbf{Landmarks:} The central dial}}\\
    \multicolumn{2}{m{0.95\columnwidth}}{\small{\textbf{Actions:} Pressing a particular sequence of buttons on the panel}}\\
    \midrule
    \small{\textbf{Test scenarios}} & {\em \textbf{Success?}} \\
    \small{Different scenes with the control panel position and incline orientation changed.} & {\em 4/5} \\
    \bottomrule
  \end{tabular}
\end{table}

Overall, our evaluation with the six tasks and 30 test scenarios showed that custom landmarks successfully enabled PbD actions to be anchored on arbitrary object or scene parts.
The robot succeeded in 26 out of the 30 total scenes it was evaluated in (Tables~\ref{tab:pbd_eval1}-\ref{tab:pbd_eval6}). 
Our search algorithm detected landmarks in cluttered or unconventional scenes, enabling PbD to be used in these arguably realistic scenarios.

Our algorithm was robust enough to avoid actions from being executed when they were not supposed to.
The robot refused to act in all cases where the necessary landmark was missing in the scene (Task 1: picking from empty rack, placing on full rack; Task 3: no bowls to pick up in scene with only cups, Task 4: no trash bin to drop in).

The failure cases were due to a few different reasons.
In Task 1, we considered one of the correct executions to be a partial success, because a Griple mount had to be manually supported to prevent it from being knocked down.
This could have been avoided if the mount (fastened with Velcro) was attached more securely to the wall.
In Task 5, our system thought that the surface of a flat box was an empty spot on the table.
This shows, again, how our system depends on the uniqueness of the custom landmark's shape to work well.
However, this could have been fixed if the robot compared the heights of the multiple empty patches it found.
In Task 2, the angle of the chair was directly lined up with the robot's Kinect sensor, causing the appearance of the chair to change too much to localize.
Similarly, in Task 6, the control panel was inclined at such an angle that the Kinect was unable to get full depth data for the dial to be localized.
These last two failure cases show that using a single sensor and a single view of the scene can make our system brittle.

These tasks illustrate the advantageous properties of custom landmarks described in Section~\ref{subsec:custom_landmarks_repr}:
\begin{enumerate}
\item Hanging tools (Task 1), grasping chairs (Task 2), picking bowls from a shelf (Task 3), and distinguishing trash bins (Task 4) all did not involve tabletop scenes.
\item Our system worked in the hard to segment shelf scene (Task 3), where bowls and cups were placed in contact.
\item Picking the top bowl from a stack of bowls (Task 3) showed that custom landmarks can be used to find partly occluded objects.
\item Picking and placing tools with Griples (Task 1) showed that custom landmarks can generalize across objects with the same part.
\item Most of the tasks would have required task-specific perception code without custom landmarks, even if an advanced object recognition system had been available.
\item Finally, the landmarks in most of the tasks did not represent objects.
Instead they represented partial objects, parts of the scene, or even the lack of objects.
\end{enumerate}

\section{Usability test}
We also conducted a small-scale usability test of the bounding box interface, used to capture custom landmarks.
5 users were taught how to use our PbD system with custom landmarks, and were asked to program one of three tasks: putting away bowls on a table, retrieving one of three snacks, or stacking cylinders on top of each other.
Participants created custom landmarks by moving the 6 sides of a 3D bounding box, as depicted in Fig.~\ref{fig:pbd}.
After completing the task, users were asked in a survey to describe the most difficult part of specifying custom landmarks.
We also recorded the time spent using the interface.
Users were recruited from email lists associated with the University of Washington.
Participants were required to not be robotics researchers and to have two years of programming experience.
All participants were male and had an average age of 25.2 years.

Participants took an average of 1 minute 11 seconds (SD=42 seconds) to specify a custom landmark.
All of the participants were able to use the bounding box interface to create viable landmarks for the corresponding PbD action.
After completing the usability test, 2 of the 5 participants wrote that they had difficulty with the interface.
One of these participants wrote that they wanted to have separate controls to translate the box, instead of adjusting the six sides of the box individually.
The other of these participants wrote that they would like to have an automatic ``guess" of the box, which could then be adjusted.
The usability test suggests that, while the bounding box interface could be improved, it is functional enough for users to work with.
 

\section{Conclusion}

This paper presents a way to enable end-users to customize the perception system of their robots, so as to support actions they would like to perform.
We proposed representing task-relevant objects, object parts, or scene parts ({\em i.e.}, landmarks) in the robot's environment with point cloud patches that are specified by the user prior to programming an action. 
The user can then program manipulation actions by demonstrating end-effector poses relative to these landmarks.
We presented an algorithm for localizing landmarks in new scenes.
We systematically evaluated our algorithm with scene-landmark pairs and characterized its performance.
Then, we showed real examples of complex manipulation tasks being programmed on a PR2 robot, and demonstrated that they work well in cluttered and unconventional (non-tabletop) scenes.
Finally, we performed a usability test demonstrating the feasibility of our system for novice users.

  \addtolength{\textheight}{-0cm} 

  \bibliographystyle{IEEEtran}
  \bibliography{landmarks}
\end{document}